\documentclass[11pt]{article}

\usepackage[preprint]{acl}

\usepackage{times}
\usepackage{latexsym}

\usepackage[T1]{fontenc}

\usepackage[utf8]{inputenc}

\usepackage{microtype}

\usepackage{inconsolata}

\usepackage{graphicx}

\usepackage{amsmath}
\usepackage{amssymb}
\usepackage{graphicx}
\usepackage{multirow}
\usepackage{booktabs}
\usepackage{svg}
\title{GAGPO: Generalized Advantage Grouped Policy Optimization}

\author{
 \textbf{Siyuan Zhu\textsuperscript{1,2}},
 \textbf{Chao Yu\textsuperscript{1}\thanks{Corresponding author.}},
 \textbf{Rongxin Yang\textsuperscript{1,2}},
 \textbf{Zongkai Liu\textsuperscript{1}},
\\
 \textbf{Jinjun Hu\textsuperscript{2}},
 \textbf{Qiwen Chen\textsuperscript{2}},
 \textbf{Yibo Zhang\textsuperscript{2}}
\\
\\
\textsuperscript{1}School of Computer Science and Engineering, Sun Yat-sen University
\textsuperscript{2}Meituan
\\
 \texttt{zhusy58@mail2.sysu.edu.cn, yuchao3@mail.sysu.edu.cn,}
\\
\texttt{zhangyibo06@meituan.com}
}

\begin{document}
\maketitle

\begin{abstract}
Reinforcement learning (RL) has emerged as a powerful paradigm for post-training large language model (LLM) agents. However, credit assignment in multi-turn environments remains a challenge. Agents typically receive sparse, trajectory-level rewards only at the end of an episode, making it difficult to identify which specific intermediate actions led to success or failure. Consequently, effectively propagating delayed outcomes back to individual steps—without relying on costly auxiliary value models—remains an open problem. 
In this paper, we propose \textbf{Generalized Advantage Grouped Policy Optimization (GAGPO)}, a critic-free RL method that enables precise, step-aligned temporal credit assignment. GAGPO constructs a non-parametric grouped value proxy from sampled rollouts to compute TD/GAE-style temporal advantages, recursively propagating outcome supervision backward through time. Coupled with group-wise advantage normalization and an action-level importance ratio, GAGPO extracts stable and localized optimization signals directly from multi-turn trajectories.
Experiments on ALFWorld and WebShop demonstrate that GAGPO outperforms strong RL baselines. Further analyses reveal faster early-stage learning, improved interaction efficiency, and smoother optimization dynamics, offering a simple yet highly effective framework for multi-turn agentic RL.
\end{abstract}

\section{Introduction}

Large language models (LLMs) are increasingly evolving from single-turn assistants into agents that can perceive environments, reason over observations, and act through multi-turn interactions~\citep{singh2025openaigpt5card, comanici2025gemini25pushingfrontier, yang2025qwen3technicalreport}. Reinforcement learning (RL)~\citep{ouyang2022traininglanguagemodelsfollow} has become a natural post-training paradigm for this transition. From PPO~\citep{schulman2017proximalpolicyoptimizationalgorithms} to critic-free grouped policy optimization methods such as GRPO~\citep{shao2024deepseekmathpushinglimitsmathematical} and its variants~\citep{ahmadian2024basicsrevisitingreinforcestyle, yu2025dapoopensourcellmreinforcement, zheng2025groupsequencepolicyoptimization, gao2025softadaptivepolicyoptimization}, online policy optimization has shown strong performance in reasoning-oriented post-training. More recently, these methods have been extended to multi-turn agent settings, enabling LLMs to improve through search, tool use, and environment interaction~\citep{wang2025ragenunderstandingselfevolutionllm, jin2025searchr1trainingllmsreason, chen2025reinforcementlearninglonghorizoninteractive}.

Despite this progress, agentic RL in multi-turn environments remains challenging: rewards are sparse and delayed, while policy optimization is typically performed at the token level, whereas task success is determined by higher-level agent actions. Consequently, intermediate decisions receive weak, noisy, and poorly localized supervision~\citep{feng2025groupingrouppolicyoptimizationllm, li2026turnppoturnleveladvantageestimation}.

Existing approaches only partially address this mismatch. One line of work introduces auxiliary critics, value estimators, or process reward models for denser step-level feedback~\citep{xi2025agentprmprocessrewardmodels, liu2025agenticreinforcementlearningimplicit, li2026turnppoturnleveladvantageestimation, li2026stabilizingoffpolicytraininglonghorizon, wei2025reinforcingmultiturnreasoningllm}, at the cost of additional training complexity and estimation error. Critic-free alternatives instead rely on trajectory-relative or Monte Carlo-style grouped optimization~\citep{feng2025groupingrouppolicyoptimizationllm, he2026hierarchyofgroupspolicyoptimizationlonghorizon}, which preserves architectural simplicity but yields high-variance, weakly propagated supervision, or on tree-structured rollouts with branch-level comparison and turn-wise reward propagation~\citep{ding2025treegrpotreeadvantagegrpoonline, zong2026at2poagenticturnbasedpolicy, dong2025agenticreinforcedpolicyoptimization}. Despite these advances, agentic RL still lacks a simple critic-free method that performs temporally propagated, step-aligned credit assignment under standard multi-turn rollouts, without auxiliary critics or specialized search procedures.

\begin{figure*}
    \centering
    \includegraphics[width=1.0\linewidth]{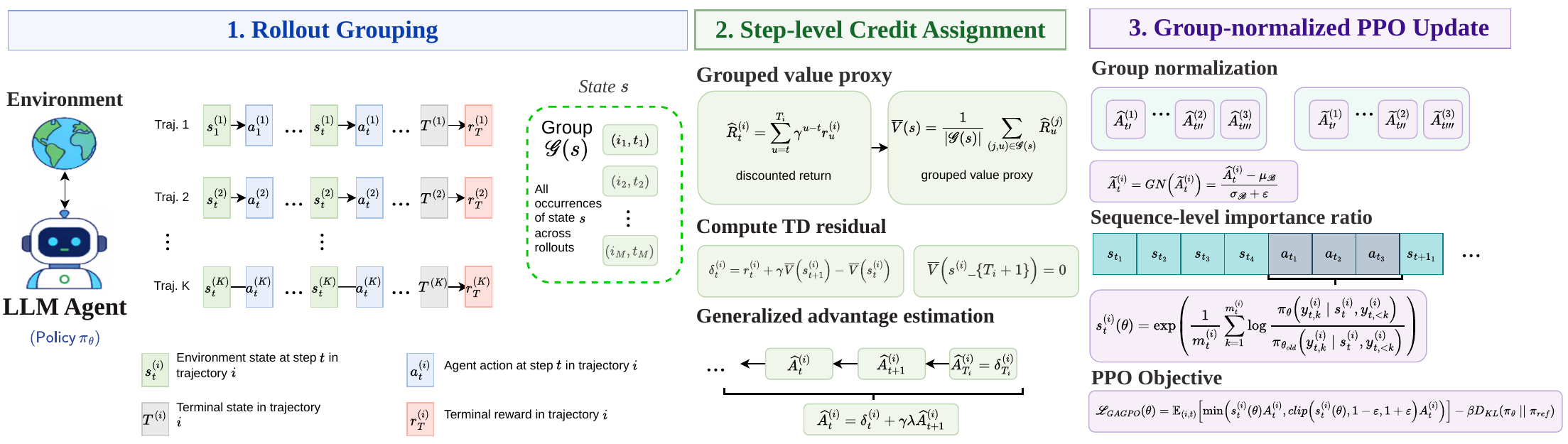}
    \caption{Overview of GAGPO.
GAGPO consists of three stages: (1) rollout grouping, which groups all occurrences of the same environment state across sampled trajectories; (2) step-level credit assignment, which builds a grouped non-parametric value proxy and computes TD/GAE-style step advantages without a learned critic; and (3) group-normalized PPO update, which normalizes step advantages within each rollout group and performs action-level policy optimization with a shared sequence-level importance ratio.}
    \label{fig:overview}
\end{figure*}

In this paper, we propose \textbf{Generalized Advantage Grouped Policy Optimization} (GAGPO), a critic-free reinforcement learning method for multi-turn agent training. GAGPO treats each environment step, rather than each token, as the basic unit of credit assignment, and constructs a non-parametric grouped value proxy from rollout groups to compute TD/GAE-style~\citep{schulman2018highdimensionalcontinuouscontrolusing} temporal advantages without learning a critic. Unlike methods that broadcast a shared trajectory-level reward to every step, GAGPO propagates outcome supervision through temporal recursion and applies group-wise advantage normalization for stability.

We evaluate GAGPO on ALFWorld~\citep{shridhar2021alfworldaligningtextembodied} and WebShop~\citep{yao2023webshopscalablerealworldweb} using Qwen2.5-1.5B-Instruct and Qwen2.5-7B-Instruct~\citep{qwen2025qwen25technicalreport}. Across both benchmarks and both model scales, GAGPO consistently outperforms strong prompting baselines and RL baselines including PPO, RLOO, GRPO, and GiGPO. Further analyses show faster early-stage learning, improved interaction efficiency, smoother optimization dynamics, and lower-variance step-level advantage signals. These results show that critic-free grouped RL can be extended more effectively to interactive LLM agents when credit is assigned at the level of environment steps and propagated through time.

\section{Background}
\subsection{Related Works}

\paragraph{RL for large language models.}
RL has become a standard paradigm for post-training LLMs. Classical RLHF~\citep{ouyang2022traininglanguagemodelsfollow} relies on PPO~\citep{schulman2017proximalpolicyoptimizationalgorithms} with a learned critic, which is costly and sensitive to value estimation, while preference-based methods such as DPO~\citep{rafailov2024directpreferenceoptimizationlanguage} bypass online RL but do not handle exploration or multi-turn interactions. Recent critic-free on-policy methods address these issues with grouped or REINFORCE-style updates, including RLOO~\citep{ahmadian2024basicsrevisitingreinforcestyle}, GRPO~\citep{shao2024deepseekmathpushinglimitsmathematical}, DAPO~\citep{yu2025dapoopensourcellmreinforcement}, and GSPO~\citep{zheng2025groupsequencepolicyoptimization}. However, these methods are designed for single-turn generation or sequence-level reasoning. GAGPO extends critic-free grouped RL to multi-turn agent training with temporally propagated, step-aligned credit assignment.

\paragraph{Credit assignment for agentic RL.}
Existing agentic RL methods address credit assignment along two directions. The first introduces auxiliary critics or process reward models for denser step-level supervision, e.g., AgentPRM~\citep{xi2025agentprmprocessrewardmodels}, iStar~\citep{liu2025agenticreinforcementlearningimplicit}, Turn-PPO~\citep{li2026turnppoturnleveladvantageestimation}, and SORL~\citep{li2026stabilizingoffpolicytraininglonghorizon}, but requires extra value or reward modeling. The second pursues finer-grained credit within critic-free grouped optimization, including anchor-state grouping in GiGPO~\citep{feng2025groupingrouppolicyoptimizationllm} and tree- or turn-structured rollouts such as Tree-GRPO~\citep{ding2025treegrpotreeadvantagegrpoonline}, AT$^2$PO~\citep{zong2026at2poagenticturnbasedpolicy}, and ARPO~\citep{dong2025agenticreinforcedpolicyoptimization}. In contrast, GAGPO stays critic-free and rollout-based, but replaces Monte Carlo or relative-return estimation with a bootstrapped TD/GAE-style temporal estimator, enabling step-aligned credit propagation without an additional critic.

\subsection{Preliminary}

\paragraph{Problem setup.}

We consider the problem of training an LLM agent to accomplish tasks through multi-turn interaction with an external environment. The interaction process is modeled as a Markov Decision Process (MDP) $\mathcal{M} = (\mathcal{S}, \mathcal{A}, P, r, \gamma)$, where $\mathcal{S}$ denotes the state space, $\mathcal{A}$ the action space, $P(s_{t+1}\mid s_t,a_t)$ the transition dynamics, $r$ the reward function, and $\gamma \in [0,1]$ the discount factor. At each step $t = 1, \dots, T$, the agent receives the environment state $s_t \in \mathcal{S}$ and generates an action $a_t \in \mathcal{A} \subseteq \mathcal{V}^{n}$, where $\mathcal{V}$ is the token vocabulary and $n$ is the maximum action length. The agent policy is parameterized by $\theta$ as $\pi_\theta(a_t \mid s_t)$. After executing $a_t$, the environment returns the next state $s_{t+1} \sim P(\cdot \mid s_t, a_t)$, where $s_{t+1}$ corresponds to the environment response represented by the updated interaction context, yielding a trajectory $\tau = \{(s_1,a_1), \dots, (s_T,a_T)\}$. Under the sparse delayed-reward setting widely studied in agentic RL, this interaction process becomes a sequential decision-making problem with challenging credit assignment.

\paragraph{Generalized advantage estimation.}
Policy optimization is commonly based on advantages assigned to sampled actions. A standard estimator is generalized advantage estimation (GAE)~\citep{schulman2018highdimensionalcontinuouscontrolusing}, which defines the TD residual $\delta_t = r_t + \gamma V(s_{t+1}) - V(s_t)$ and computes
\[
\hat{A}_t = \sum_{l=0}^{T-t} (\gamma\lambda)^l \delta_{t+l},
\]
where $V(\cdot)$ is a value function and $\lambda \in [0,1]$ controls the bias--variance trade-off. By recursively propagating TD residuals backward through time, GAE provides a temporally structured credit signal, but relies on a learned value function that is absent in critic-free grouped policy optimization.

\begin{table*}[t]
\centering
\caption{Performance on ALFWorld and WebShop.}
\label{tab:main_results}
\resizebox{\textwidth}{!}{
\begin{tabular}{llccccccc|cc}
\toprule
\multirow{2}{*}{Type} & \multirow{2}{*}{Method} & \multicolumn{7}{c|}{\textbf{ALFWorld}} & \multicolumn{2}{c}{\textbf{WebShop}} \\
 & & Pick & Look & Clean & Heat & Cool & Pick2 & All & Score & Succ.\\
\midrule
\multicolumn{10}{l}{\textit{Closed-Source Model}} \\
Prompting& GPT-4o & 75.3 & 60.8 & 31.2 & 56.7 & 21.6 & 49.8 & 48.0& 31.8 & 23.7\\
Prompting& Gemini-2.5-Pro & 92.8 & 63.3 & 62.1 & 69.0 & 26.6 & 58.7 & 60.3& 42.5 & 35.9\\
\midrule
\multicolumn{10}{l}{\textit{Qwen2.5-1.5B-Instruct}} \\
Prompting& Qwen2.5 & 5.9 & 5.5 & 3.3 & 9.7 & 4.2 & 0.0 & 4.1 & 23.1 & 5.2\\
Prompting& ReAct & 17.4 & 20.5 & 15.7 & 6.2 & 7.7 & 2.0 & 12.8& 40.1& 11.3\\
Prompting& Reflexion & 35.3 & 22.2 & 21.7 & 13.6 & 19.4 & 3.7 & 21.8 & 55.8& 21.9\\
RL Training& PPO (with critic) & 64.8\textsubscript{\textpm3.5} & 40.5\textsubscript{\textpm6.9} & 57.1\textsubscript{\textpm4.9} & 60.6\textsubscript{\textpm6.6} & 46.4\textsubscript{\textpm4.0} & 47.4\textsubscript{\textpm1.9} & 54.4\textsubscript{\textpm3.1}& 73.8\textsubscript{\textpm3.0} & 51.5\textsubscript{\textpm2.9} \\
RL Training& RLOO & 88.3\textsubscript{\textpm3.0} & 52.8\textsubscript{\textpm8.6} & 71.0\textsubscript{\textpm5.9} & 62.8\textsubscript{\textpm8.7} & 66.4\textsubscript{\textpm5.5} & 56.9\textsubscript{\textpm4.7} & 69.7\textsubscript{\textpm2.5}& 73.9\textsubscript{\textpm5.6}& 52.1\textsubscript{\textpm6.7}\\
RL Training& GRPO & 73.1\textsubscript{\textpm3.4} & 66.7\textsubscript{\textpm10.1} & 80.2\textsubscript{\textpm8.2} & 69.6\textsubscript{\textpm12.2} & 58.7\textsubscript{\textpm4.5} & 67.6\textsubscript{\textpm11.0} & 
70.3\textsubscript{\textpm3.6} &
80.5\textsubscript{\textpm2.0}&
66.4\textsubscript{\textpm4.4}
\\
RL Training& GiGPO & 98.4\textsubscript{\textpm2.1} & 
72.2\textsubscript{\textpm4.9} & 
91.1\textsubscript{\textpm6.1} & 
\textbf{96.8}\textsubscript{\textpm6.25} & 
82.6\textsubscript{\textpm4.5} & 
79.7\textsubscript{\textpm5.4} & 
88.1\textsubscript{\textpm1.95} & 
79.8\textsubscript{\textpm1.2}&
62.5\textsubscript{\textpm1.1}\\

RL Training & \textbf{GAGPO (Ours) }
& \textbf{99.2}\textsubscript{\textpm3.1} 
& \textbf{83.8}\textsubscript{\textpm6.3} 
& \textbf{97.3}\textsubscript{\textpm1.9} 
& 95.1\textsubscript{\textpm3.5} 
& \textbf{84.9}\textsubscript{\textpm1.8} 
& \textbf{89.8}\textsubscript{\textpm6.0} 
& \textbf{93.5}\textsubscript{\textpm1.3}
& \textbf{88.6}\textsubscript{\textpm3.3} 
& \textbf{78.1}\textsubscript{\textpm1.1}\\

\midrule
\multicolumn{10}{l}{\textit{Qwen2.5-7B-Instruct}} \\
Prompting& Qwen2.5 & 33.4 & 21.6 & 19.3 & 6.9 & 2.8 & 3.2 & 14.8 & 26.4 & 7.8\\
Prompting& ReAct & 48.5 & 35.4 & 34.3 & 13.2 & 18.2 & 17.6 & 31.2 & 46.2 & 19.5\\
Prompting& Reflexion & 62.0 & 41.6 & 44.9 & 30.9 & 36.3 & 23.8 & 42.7& 58.1& 28.8\\
RL Training& PPO (with critic) & 92.3\textsubscript{\textpm4.0} & 64.0\textsubscript{\textpm8.4} & 92.5\textsubscript{\textpm2.4} & 89.5\textsubscript{\textpm7.0} & 80.3\textsubscript{\textpm2.0} & 68.8\textsubscript{\textpm8.3} & 80.4\textsubscript{\textpm2.7} & 81.4\textsubscript{\textpm3.1}& 68.7\textsubscript{\textpm5.1}\\
RL Training& RLOO & 87.6\textsubscript{\textpm4.3} & 78.2\textsubscript{\textpm8.3} & 87.3\textsubscript{\textpm5.8} & 81.3\textsubscript{\textpm7.6} & 71.9\textsubscript{\textpm5.2} & 48.9\textsubscript{\textpm8.4} & 75.5\textsubscript{\textpm4.6} & 80.3\textsubscript{\textpm3.2} & 65.7\textsubscript{\textpm4.0}\\
RL Training& GRPO & 85.9\textsubscript{\textpm6.9} & 
69.5\textsubscript{\textpm4.8} & 
82.7\textsubscript{\textpm6.6} & 
73.7\textsubscript{\textpm6.8} & 
65.4\textsubscript{\textpm8.4} & 
62.6\textsubscript{\textpm6.3} & 
73.2\textsubscript{\textpm4.6} & 
80.5\textsubscript{\textpm2.1} & 
66.8\textsubscript{\textpm1.7}\\ 
RL Training& GiGPO & 
96.2\textsubscript{\textpm3.9} &
90.9\textsubscript{\textpm9.1} & 
95.5\textsubscript{\textpm5.1} & 
80.9\textsubscript{\textpm8.7} & 
72.1\textsubscript{\textpm8.6} & 
90.4\textsubscript{\textpm5.1} & 
88.8\textsubscript{\textpm4.5} & 
86.3\textsubscript{\textpm2.7} & 
73.3\textsubscript{\textpm1.9}\\

RL Training & \textbf{GAGPO (Ours)}
& \textbf{97.8}\textsubscript{\textpm1.6} 
& \textbf{97.8}\textsubscript{\textpm3.1} 
& \textbf{95.8}\textsubscript{\textpm5.9} 
& \textbf{97.6}\textsubscript{\textpm3.3} 
& \textbf{92.1}\textsubscript{\textpm3.0} 
& \textbf{92.6}\textsubscript{\textpm5.4} 
& \textbf{95.6}\textsubscript{\textpm0.9}
& \textbf{90.3}\textsubscript{\textpm1.2} 
& \textbf{77.5}\textsubscript{\textpm3.0}\\

\bottomrule
\end{tabular}

}
\end{table*}

\begin{figure*}[t]
    \centering
    \includegraphics[width=1.0\linewidth]{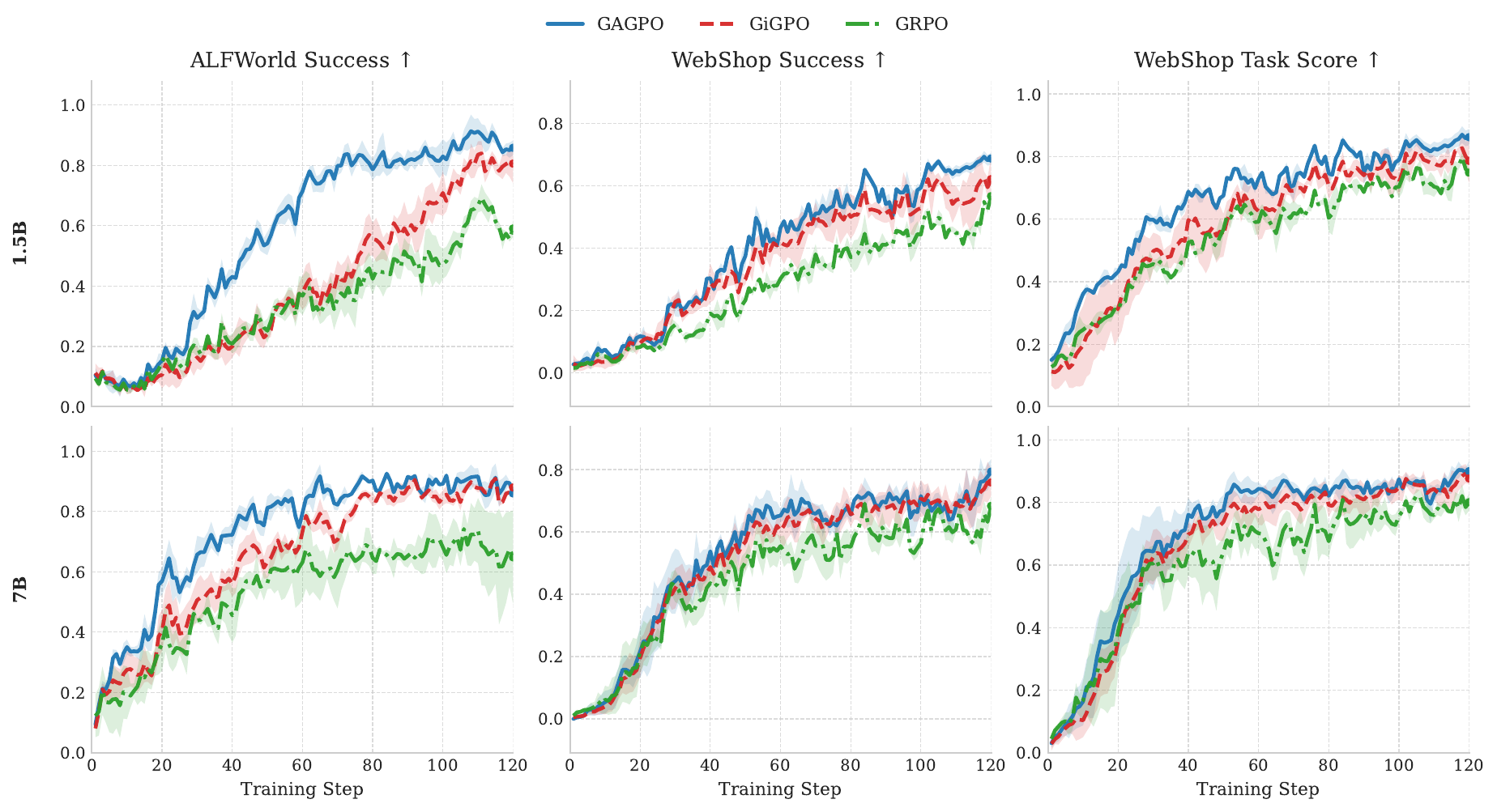}
    \caption{
    Learning dynamics on ALFWorld and WebShop over the first 120 training steps for Qwen2.5-1.5B-Instruct and Qwen2.5-7B-Instruct.
    The figure reports ALFWorld success rate, WebShop success rate, and WebShop task score.
    Across both backbones, GAGPO improves faster than GiGPO and GRPO in the early stage of training and maintains stronger overall performance throughout most of training.
    }
    \label{fig:learning_main}
\end{figure*}

\section{Method}

\label{sec:method}

\textbf{Generalized Advantage Grouped Policy Optimization} (GAGPO) is a critic-free RL algorithm for multi-turn agentic training (Figure~\ref{fig:overview}). Building on the PPO-style grouped optimization framework, GAGPO replaces direct Monte Carlo-style relative advantages with a temporally propagated step-level estimator, and uses a shared sequence-level importance ratio aligned with the action boundary rather than individual tokens. The key idea is to construct a non-parametric value proxy from grouped rollouts and compute TD/GAE-style advantages over environment steps without an additional critic. This design provides (i) \emph{step alignment} with the agent's decision boundary, (ii) \emph{temporal credit propagation} of delayed outcomes, and (iii) \emph{critic-free bootstrapping}.

Formally, for a given task instance, a rollout group $\mathcal{T}=\{\tau^{(i)}\}_{i=1}^{K}$, where $\tau^{(i)}=\{(s^{(i)}_t,a^{(i)}_t,r^{(i)}_t)\}_{t=1}^{T_i}$ and each action $a^{(i)}_t = (y^{(i)}_{t,1},\dots,y^{(i)}_{t,m^{(i)}_t})$ is a token sequence.

\subsection{Step-Aligned Grouped Temporal Credit Assignment}
\label{subsec:temporal_adv}

Since rewards are sparse and delayed while policy updates operate at the token level, GAGPO treats each \emph{environment step} as the unit of credit assignment: all tokens within the same action $a^{(i)}_t$ share a single step-level advantage $\hat{A}^{(i)}_t$.

To construct critic-free temporal credit signals, GAGPO organizes rollout steps into state-consistent groups. For each state $s$, the corresponding step group $\mathcal{G}(s)=\{(i,t)\mid s^{(i)}_t=s\}$ gathers all occurrences of $s$ across the rollout group, built entirely from collected trajectories at no extra rollout cost.

For each sampled step $(i,t)$, its discounted return is defined as
\[
    \hat{R}^{(i)}_t = \sum_{u=t}^{T_i}\gamma^{\,u-t} r^{(i)}_u,
\]
where $\gamma\in[0,1]$ is the discount factor. A non-parametric grouped value proxy for state $s$ is constructed by averaging the discounted returns of steps in the same group:
\[
    \bar{V}(s)=\frac{1}{|\mathcal{G}(s)|}\sum_{(j,u)\in \mathcal{G}(s)} \hat{R}^{(j)}_u.
\]
Based on this grouped value proxy, GAGPO computes a temporal-difference residual at each step:
\[
    \delta^{(i)}_t = r^{(i)}_t + \gamma \bar{V}(s^{(i)}_{t+1}) - \bar{V}(s^{(i)}_t),
\]
where $\bar{V}(s^{(i)}_{T_i+1})=0$ for terminal states. The step-level temporal advantage is then defined recursively in a GAE-style manner:
\begin{equation}
    \hat{A}^{(i)}_t = \delta^{(i)}_t + \gamma \lambda \hat{A}^{(i)}_{t+1},
    \label{eq:gae_recursion}
\end{equation}
where $\lambda\in[0,1]$ controls the bias--variance trade-off in temporal credit propagation. Equivalently,
\[
    \hat{A}^{(i)}_t
    =
    \sum_{l=0}^{T_i-t}
    (\gamma\lambda)^l \delta^{(i)}_{t+l}.
\]

\subsection{Localized Objective and Group-Normalized PPO Optimization}
\label{subsec:ppo_objective}

Many grouped policy optimization methods combine local step-level signals with a trajectory-level reward or relative advantage. However, adding the same episode-level offset to every step makes all actions share an identical global component regardless of their temporal position, reducing contrast among intermediate decisions. GAGPO instead uses the temporal advantage in Eq.~\ref{eq:gae_recursion} as the sole optimization signal: episode-level outcomes still influence earlier decisions through temporal recursion, without imposing a uniform offset on all steps.

Although the temporal estimator improves credit localization, step advantage magnitudes still vary across tasks and rollout groups. Since batch-level normalization mixes heterogeneous tasks and disrupts the within-group structure, GAGPO applies \emph{group normalization}: let $\mathcal{B}$ denote all sampled steps in the same rollout group; $\hat{A}^{(i)}_t$ is standardized as ${A}_t^{(i)} = (\hat{A}^{(i)}_t-\mu_{\mathcal{B}})/(\sigma_{\mathcal{B}}+\epsilon)$, where $\mu_{\mathcal{B}},\sigma_{\mathcal{B}}$ are group statistics, preserving within-group comparisons while mitigating cross-task scale variation.

Finally, GAGPO optimizes the policy with a PPO-style clipped objective. Since each action $a^{(i)}_t$ is a sequence of tokens, similar to~\citet{zheng2025groupsequencepolicyoptimization}, the same normalized step-level advantage $A^{(i)}_t$ is assigned to all tokens within that action. Rather than clipping token-wise importance ratios independently, a length-normalized ratio is computed for each action sequence by averaging token-level log-ratios within the action and exponentiating:
\[
    s^{(i)}_t(\theta)
    =
    \exp\!\left(
    \frac{1}{m^{(i)}_t}
    \sum_{k=1}^{m^{(i)}_t}
    \log
    \frac{
    \pi_{\theta}(y^{(i)}_{t,k}\mid s^{(i)}_t, y^{(i)}_{t,<k})
    }{
    \pi_{\theta_{\mathrm{old}}}(y^{(i)}_{t,k}\mid s^{(i)}_t, y^{(i)}_{t,<k})
    }
    \right),
\]
where $m^{(i)}_t$ is the number of valid tokens in action $a^{(i)}_t$. This defines a sequence-level ratio for the entire action, while normalizing for action length.

The clipped objective is then written as
\[
\begin{aligned}
\mathcal{L}_{\mathrm{GAGPO}}(\theta)
&= \mathbb{E}_{(i,t)} \Big[
\min \Big(
    s^{(i)}_t(\theta){A}^{(i)}_t, \\
&\qquad\qquad
    \mathrm{clip}\!\big(s^{(i)}_t(\theta), 1-\epsilon, 1+\epsilon\big){A}^{(i)}_t
\Big)
\Big] \\
&\quad - \beta\, D_{\mathrm{KL}}\!\left(\pi_\theta \,\|\, \pi_{\mathrm{ref}}\right),
\end{aligned}
\]
where $\epsilon$ is the PPO clipping coefficient, $\beta$ controls the KL penalty strength, and $\pi_{\mathrm{ref}}$ is a reference policy.

Overall, GAGPO preserves the simplicity and efficiency of grouped policy optimization while introducing a temporally propagated and step-aligned credit signal for multi-turn agent training.

\begin{figure}[t]
    \centering
    \includegraphics[width=1.0\linewidth]{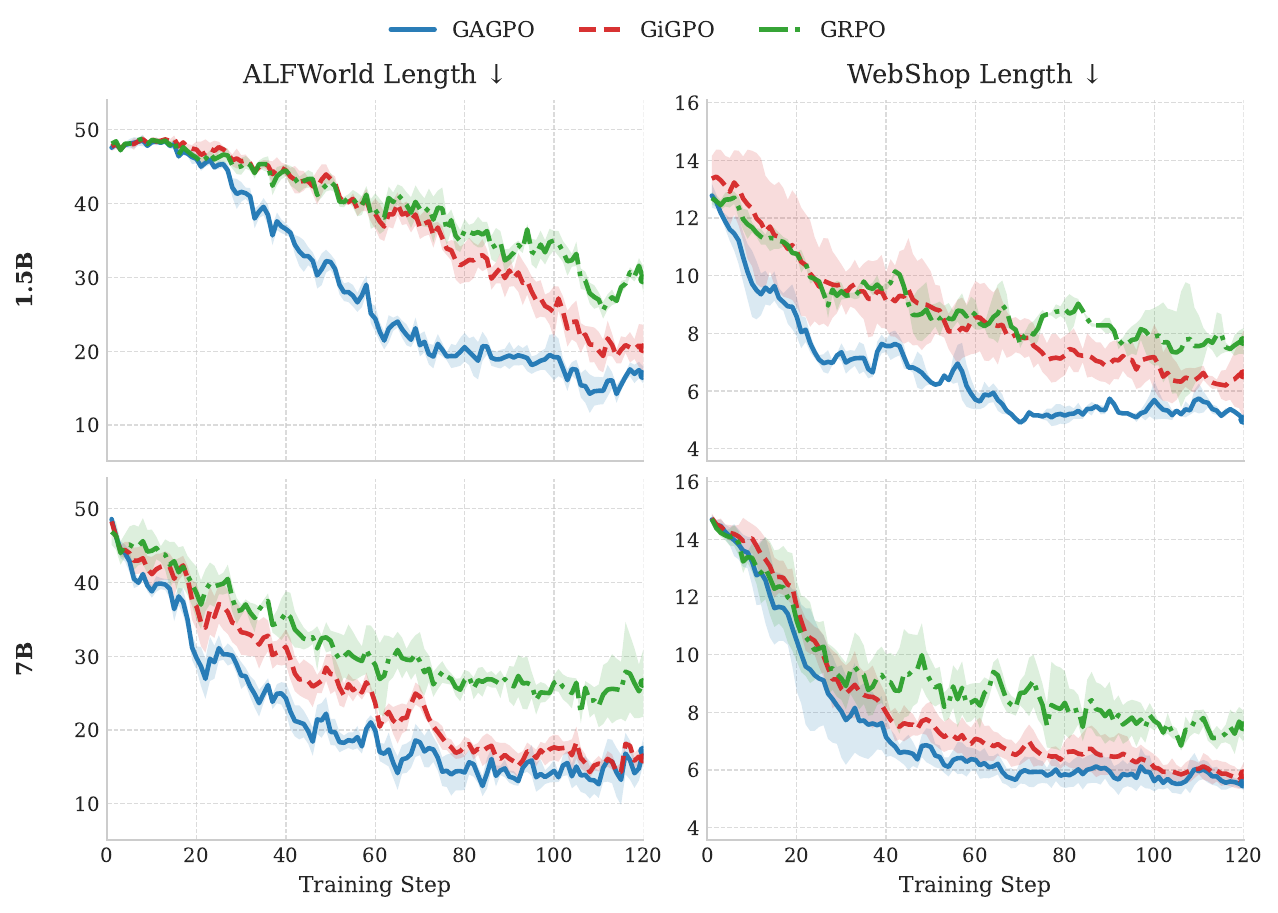}
    \caption{
    Average episode length on ALFWorld and WebShop over the first 120 training steps for Qwen2.5-1.5B-Instruct and Qwen2.5-7B-Instruct.
    }
    \label{fig:learning_length}
\end{figure}

\begin{figure*}[t]
    \centering
    \includegraphics[width=1\linewidth]{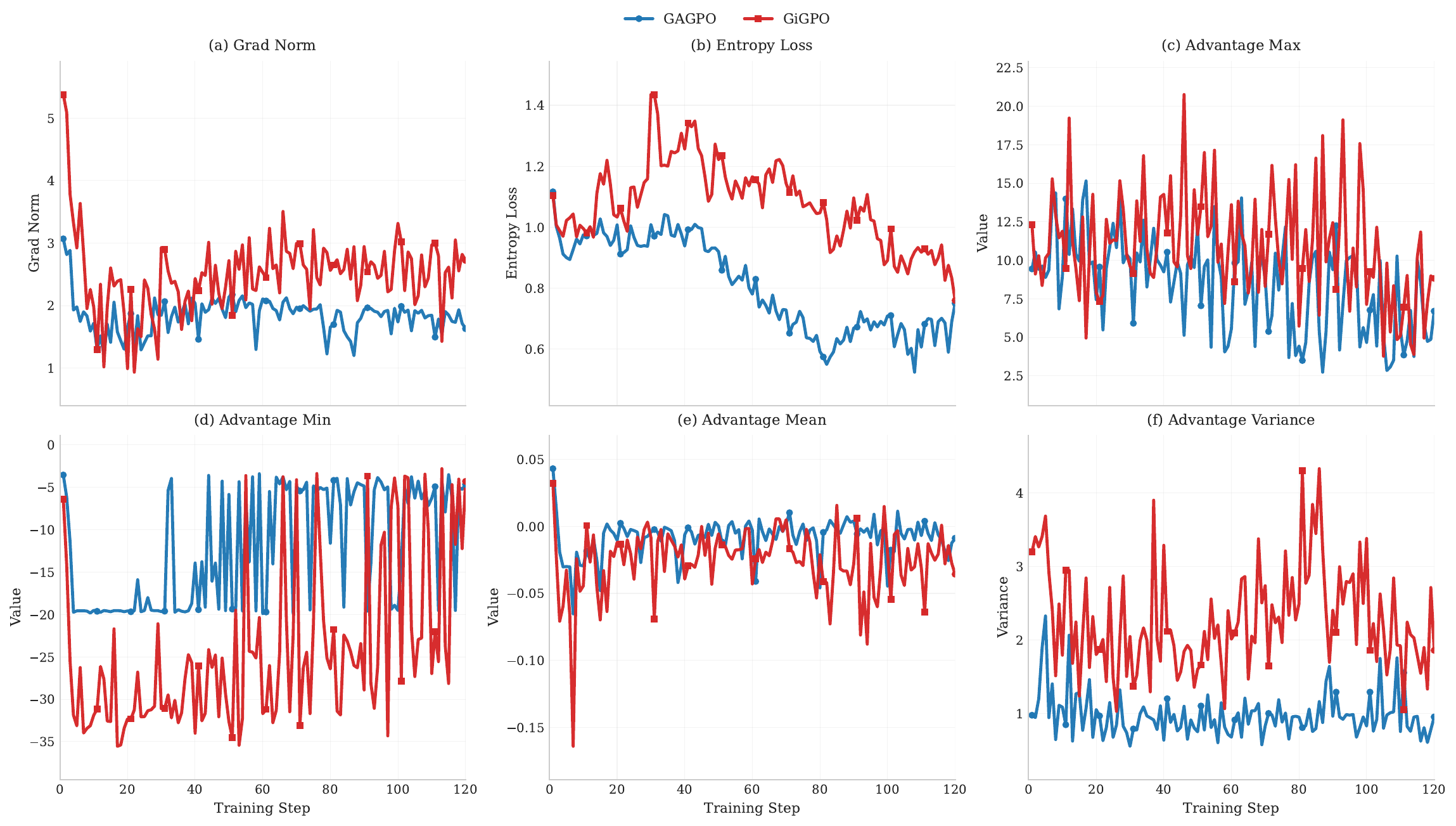}
    \caption{
    Optimization and advantage statistics of GAGPO and GiGPO on ALFWorld over the first 120 training steps, including gradient norm, entropy loss, and summary statistics of step-level advantages.
    Compared with GiGPO, GAGPO exhibits smoother gradient dynamics, faster entropy reduction, lower advantage variance, and substantially tighter advantage extrema, indicating more stable optimization and lower-variance credit signals.
    }
    \label{fig:alf_metrics_15b}
\end{figure*}

\begin{figure}[t]
    \centering
    \includegraphics[width=1\linewidth]{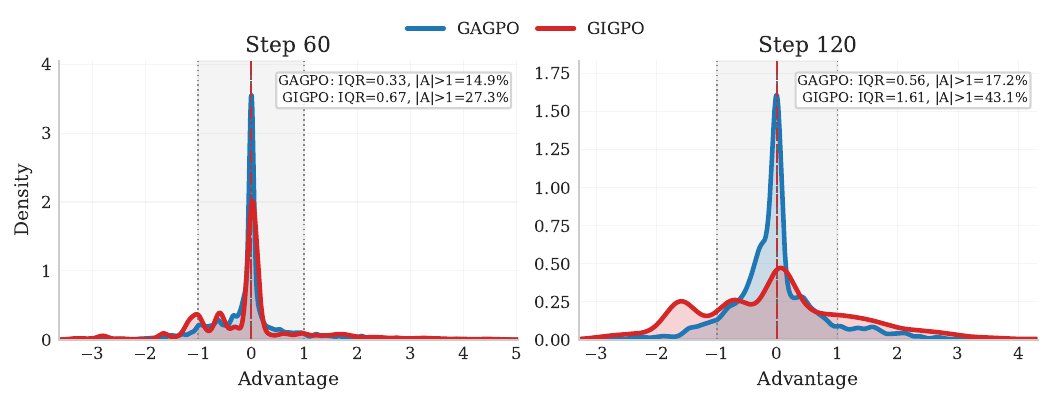}
    \caption{
    Distribution of normalized step-level advantages at training steps 60 and 120 on ALFWorld.
    The gray region marks $[-1,1]$, and the inset reports the interquartile range (IQR) and the fraction of large-magnitude advantages with $|A|>1$.
    GAGPO shows smaller spread and lower tail mass than GiGPO at both stages.
    }
    \label{fig:adv_dist}
\end{figure}

\section{Experiments}

\subsection{Experimental Setup}

\paragraph{Benchmarks.}
We evaluate GAGPO on two representative multi-turn agent benchmarks, \textbf{ALFWorld}~\citep{shridhar2021alfworldaligningtextembodied} and \textbf{WebShop}~\citep{yao2023webshopscalablerealworldweb}. ALFWorld requires sequential decision making over embodied household tasks such as finding, manipulating, and composing objects, while WebShop evaluates interactive decision making in an online shopping environment via multi-turn search, comparison, and selection. Both environments are purely text-based with structured, deterministic observations, allowing us to aggregate same-state occurrences via exact textual match when constructing the grouped value proxy $\bar{V}(s)$. Episodes terminate upon task success or upon reaching a fixed interaction budget (50 steps for ALFWorld, 15 for WebShop). Following prior work~\citep{feng2025groupingrouppolicyoptimizationllm, he2026hierarchyofgroupspolicyoptimizationlonghorizon}. The evaluation reports category-wise success rates and overall average success on ALFWorld, and the average score and success rate on WebShop.

\paragraph{Baselines.}

GAGPO is compared against both prompting-based and RL-based baselines. Prompting baselines include direct prompting, ReAct~\citep{yao2023reactsynergizingreasoningacting}, and Reflexion~\citep{shinn2023reflexionlanguageagentsverbal} on Qwen2.5 backbones, as well as strong closed-source models such as GPT-4o and Gemini-2.5-Pro. RL baselines include PPO~\citep{ouyang2022traininglanguagemodelsfollow} with a learned critic, RLOO~\citep{ahmadian2024basicsrevisitingreinforcestyle}, GRPO~\citep{shao2024deepseekmathpushinglimitsmathematical}, and GiGPO~\citep{feng2025groupingrouppolicyoptimizationllm}.
For prompting baselines, PPO, and RLOO, we follow the results reported by GiGPO under the same backbones, environments, and evaluation protocols. For GRPO and GiGPO, we re-run the baselines under the same training and evaluation pipeline as GAGPO to ensure a controlled comparison. The implementation follows GiGPO exactly in all training and evaluation settings, except for the proposed credit assignment mechanism used by GAGPO.

\begin{table*}[t]
\centering
\caption{Ablation study on key components of GAGPO. We report ALFWorld overall success rate, WebShop average score, and WebShop success rate on both Qwen2.5-1.5B-Instruct and Qwen2.5-7B-Instruct backbones.}
\label{tab:ablation_full}

\resizebox{\textwidth}{!}{
\begin{tabular}{lccc|ccc}
\toprule
\multirow{2}{*}{Method}
& \multicolumn{3}{c|}{\textbf{Qwen2.5-1.5B-Instruct}}
& \multicolumn{3}{c}{\textbf{Qwen2.5-7B-Instruct}} \\
\cmidrule(lr){2-4} \cmidrule(lr){5-7}
& ALFWorld All & WebShop Score & WebShop Succ.
& ALFWorld All & WebShop Score & WebShop Succ. \\
\midrule

Full GAGPO
& 93.5\textsubscript{\textpm1.3}
& 88.6\textsubscript{\textpm3.3}
& 78.1\textsubscript{\textpm1.1}
& 95.6\textsubscript{\textpm0.9}
& 90.3\textsubscript{\textpm1.2}
& 77.5\textsubscript{\textpm3.0} \\
\midrule

w/o temporal recursion ($\lambda = 0$)
& 85.8\textsubscript{\textpm2.2}
& 82.5\textsubscript{\textpm2.7}
& 64.9\textsubscript{\textpm3.8}
& 89.9\textsubscript{\textpm1.9}
& 83.8\textsubscript{\textpm2.1}
& 71.5\textsubscript{\textpm2.3} \\

MC-style step advantage
& 87.1\textsubscript{\textpm1.1}
& 83.1\textsubscript{\textpm1.6}
& 65.0\textsubscript{\textpm3.2}
& 91.2\textsubscript{\textpm0.8}
& 86.4\textsubscript{\textpm2.9}
& 73.8\textsubscript{\textpm3.2} \\

w/o action sequence importance ratio
& 89.6\textsubscript{\textpm1.8}
& 86.8\textsubscript{\textpm2.5}
& 73.5\textsubscript{\textpm2.8}
& 92.8\textsubscript{\textpm1.4}
& 88.5\textsubscript{\textpm1.6}
& 74.1\textsubscript{\textpm1.9} \\

w/ trajectory reward broadcast
& 88.2\textsubscript{\textpm2.0}
& 85.3\textsubscript{\textpm2.6}
& 70.1\textsubscript{\textpm3.4}
& 92.1\textsubscript{\textpm1.5}
& 87.8\textsubscript{\textpm1.8}
& 74.2\textsubscript{\textpm2.0} \\

batch normalization
& 90.9\textsubscript{\textpm1.6}
& 88.1\textsubscript{\textpm3.0}
& 75.2\textsubscript{\textpm3.1}
& 92.3\textsubscript{\textpm1.2}
& 89.2\textsubscript{\textpm1.5}
& 76.6\textsubscript{\textpm1.7} \\

w/o normalization
& 84.7\textsubscript{\textpm2.4}
& 81.0\textsubscript{\textpm3.2}
& 62.8\textsubscript{\textpm4.1}
& 85.6\textsubscript{\textpm2.2}
& 82.4\textsubscript{\textpm2.5}
& 68.1\textsubscript{\textpm2.7} \\

\bottomrule
\end{tabular}
}
\end{table*}

\paragraph{Implementation details.}
Experiments are conducted on Qwen2.5-1.5B-Instruct and Qwen2.5-7B-Instruct. To ensure a controlled comparison, all training and evaluation settings are kept identical to GiGPO, including the rollout group size, optimizer, learning rate, batch size, mini batch size, clipping coefficient, KL regularization, and environment settings, etc. GAGPO introduces two method-specific hyperparameters: the discount factor $\gamma$ and the temporal propagation coefficient $\lambda$, which are set to 0.95 and 0.8, respectively, unless otherwise specified. We report the mean and standard deviation over 3 random seeds. Unless otherwise specified, main results in Table~\ref{tab:main_results} are reported at the final checkpoint after 160 training steps, while Figures~\ref{fig:learning_main} and~\ref{fig:learning_length} visualize the first 120 training steps for clarity.
We provide additional hyperparameter sensitivity results and exact-match group-size statistics in Appendices~\ref{sec:appendix_hparam} and~\ref{sec:appendix_group_dist}.

\subsection{Main Results}

Table~\ref{tab:main_results} reports the main comparison on ALFWorld and WebShop. GAGPO consistently outperforms all RL baselines on aggregate metrics across both benchmarks and both model scales. 
On ALFWorld, GAGPO improves the overall score from 88.1 to 93.5 on Qwen2.5-1.5B and from 88.8 to 95.6 on Qwen2.5-7B, achieving gains of 5.4 and 6.8 points over the strongest baseline, respectively. 
On WebShop, GAGPO raises the score from 80.5 to 88.6 and the success rate from 66.4 to 78.1 on the 1.5B model. On the 7B model, it further improves the score from 86.3 to 90.3 and the success rate from 73.3 to 77.5, corresponding to gains of 8.1/11.7 and 4.0/4.2 points over the strongest baseline, respectively.


\subsection{Analysis}

\paragraph{Learning dynamics and interaction efficiency.}

We further analyze the learning dynamics of GAGPO during the first 120 training steps, where differences in credit assignment are most directly reflected in optimization efficiency. For Figures~\ref{fig:learning_main} and~\ref{fig:learning_length}, each curve reports the mean over three random seeds, shaded regions denote standard deviation, and moving-average smoothing with coefficient 0.6 is applied for visualization. Each training step corresponds to one policy update based on a batch of sampled trajectories, so the curves in Figures~\ref{fig:learning_main} and~\ref{fig:learning_length} reflect early-stage policy improvement under the same training budget.

As shown in Figure~\ref{fig:learning_main}, GAGPO consistently improves faster than GiGPO and GRPO across both benchmarks and both model scales. On ALFWorld, the advantage is especially pronounced: GAGPO reaches substantially higher success rates early in training and maintains a clear gap over most of the optimization trajectory. On WebShop, the gains are more moderate but remain consistent in both success rate and task score. 
Since WebShop task score is computed from the final reward and reflects how well the selected product satisfies the instruction, including partial matches in attributes, options, type, and price, these results suggest that GAGPO improves not only exact task completion but also the quality of the final product selection.

Figure~\ref{fig:learning_length} provides a complementary view from interaction efficiency. Across both ALFWorld and WebShop, GAGPO generally achieves shorter average episode lengths as training progresses, with the difference being particularly visible on ALFWorld. Because unsuccessful episodes are truncated by the maximum interaction budget, episode length should not be interpreted in isolation. Instead, taken together with the higher success rates in Figure~\ref{fig:learning_main}, the lower episode lengths indicate that GAGPO reaches successful completion in fewer interaction steps on average.

Overall, these results suggest that GAGPO converts training signal into successful behavior more efficiently in the early stage of optimization, consistent with the claim that temporally propagated and step-aligned credit assignment provides a more effective learning signal for multi-turn agent training.

\paragraph{Optimization stability and advantage statistics.}
To better understand the source of GAGPO's gains, we analyze optimization dynamics and step-level advantage statistics on ALFWorld with Qwen2.5-1.5B-Instruct over the first 120 training steps. Since GAGPO and GiGPO share the same rollout grouping and training pipeline, and differ mainly in the step-level credit estimator, these metrics provide a direct view of whether the proposed temporal estimator yields more stable updates and lower-variance learning signals.

As shown in Figure~\ref{fig:alf_metrics_15b}, GAGPO exhibits consistently smoother optimization dynamics than GiGPO. After the initial warm-up phase, the gradient norm under GAGPO remains lower and less volatile, whereas GiGPO shows frequent high-amplitude fluctuations throughout training. The entropy loss also decreases faster and more monotonically under GAGPO, suggesting that the policy converts exploration into more confident task-specific behavior earlier in training.

The advantage statistics in Figure~\ref{fig:adv_dist} further support this trend. 
Although both methods keep the normalized mean close to zero, GAGPO produces a more concentrated distribution with substantially lower tail mass. 
At step 60, GAGPO reduces the interquartile range from 0.67 to 0.33 and the fraction of large-magnitude advantages with $|A|>1$ from 27.3\% to 14.9\% compared with GiGPO. 
The gap becomes more pronounced at step 120, where GiGPO exhibits a much broader distribution with an IQR of 1.61 and 43.1\% large-magnitude advantages, while GAGPO maintains a compact distribution with an IQR of 0.56 and only 17.2\% large-magnitude advantages. 
Together with the smoother gradient dynamics in Figure~\ref{fig:alf_metrics_15b}, these results indicate that GAGPO reduces extreme step-level credit signals rather than merely shifting the advantage mean.

This behavior is consistent with the design of the proposed estimator. 
GiGPO combines trajectory-level relative feedback with step-level Monte Carlo-style signals, which can introduce large variations when delayed outcomes are assigned to multiple intermediate actions. 
In contrast, GAGPO propagates outcome supervision through TD/GAE-style temporal recursion and applies a single group-wise normalization to the resulting step-level advantages. 
This yields a more localized and lower-variance optimization signal, reducing the chance that PPO updates are dominated by noisy high-magnitude advantages. 
Importantly, the sharper concentration around zero should not be interpreted as weakened learning signal, since GAGPO simultaneously achieves higher task performance and smoother optimization; rather, it suggests that well-learned or low-disagreement states receive smaller residual updates while informative steps still provide effective credit for policy improvement.

\subsection{Ablation Study}

We conduct ablations to examine the role of each component in GAGPO. As shown in Table~\ref{tab:ablation_full}, the full method consistently achieves the best performance on both Qwen2.5-1.5B-Instruct and Qwen2.5-7B-Instruct across ALFWorld and WebShop, showing that the gains come from the combination of temporally propagated credit assignment, localized step-level optimization, and group normalization.

Removing temporal recursion by setting $\lambda=0$ leads to clear performance drops on both benchmarks, as the truncated temporal horizon fails to propagate delayed success signals. Replacing our TD/GAE-style estimator with the MC-style step advantage performs slightly better than the myopic $\lambda=0$ variant by capturing full trajectory returns, but it still falls significantly short of the full GAGPO. This demonstrates that our GAE-style temporal propagation successfully achieves a superior bias-variance trade-off compared to both myopic (TD) and high-variance (MC) alternatives.

To evaluate the necessity of step-aligned policy updates, we ablate the shared action sequence importance ratio, thereby reverting to standard token-independent PPO clipping. This variant suffers a noticeable performance drop across both benchmarks. This decline demonstrates that when assigning a single step-level advantage to a multi-token action, treating the token importance ratios independently can cause inconsistent updates and gradient tearing within the action. By using a shared sequence-level ratio, GAGPO ensures that the entire action remains a cohesive optimization unit.

We further compare against adding a trajectory-level reward broadcast term to every step. Although this variant performs better than several weaker ablations, it remains consistently worse than the full method, suggesting that directly injecting the same episode-level offset into all steps weakens local credit assignment. In contrast, GAGPO preserves outcome supervision through temporal recursion while avoiding indiscriminate trajectory-wide bias.

Finally, normalization is crucial for stable optimization. Removing normalization causes the largest overall degradation, while replacing group normalization with standard batch normalization also hurts performance. This suggests that advantage normalization should respect the grouped rollout structure: group-wise normalization preserves meaningful within-task comparisons and improves robustness across heterogeneous trajectories.

\section{Conclusion}

We presented GAGPO, a critic-free grouped policy optimization method for multi-turn agentic RL. By aligning credit assignment with environment steps, propagating sparse outcome supervision through a TD/GAE-style temporal recursion over a non-parametric grouped value proxy, and applying group-wise normalization, GAGPO provides a step-aligned, temporally consistent, and stable training signal for LLM agents. Experiments on ALFWorld and WebShop with Qwen2.5-1.5B/7B-Instruct show consistent gains over strong RL baselines, with improved learning efficiency and optimization stability. Future work includes extending the grouping mechanism beyond exact state matches to support approximate state aggregation in partially observed environments.

\section{Limitations}

While GAGPO provides a simple and effective framework for step-aligned credit assignment in multi-turn agentic RL, several limitations remain.

\paragraph{Reliance on exact state matching.}
The non-parametric grouped value proxy \(\bar{V}(s)\) is constructed by aggregating rollouts that share the same environment state \(s\). In ALFWorld and WebShop, observations are textual and deterministic, so the same state can be identified via exact string matching. In environments with stochastic observations, continuous sensory inputs, or partial observability, exact matches become rare and \(|\mathcal{G}(s)|\) shrinks toward one, weakening the grouped value proxy and reducing GAGPO toward a per-trajectory Monte Carlo estimate. Extending GAGPO to such settings will require approximate state aggregation, such as embedding-based clustering or learned equivalence relations.

\paragraph{Scope of evaluation and assumptions.}
Our experiments focus on sparse episode-end rewards, discrete text-based actions, and two representative benchmarks, ALFWorld and WebShop, with Qwen2.5-1.5B/7B-Instruct backbones. We do not study settings with dense process rewards, mixed reward sources, continuous, asynchronous interaction, or substantially larger and more diverse agent domains. Further validation is needed to determine whether the same temporal estimator remains beneficial in richer long-horizon environments.

\paragraph{Potential risks.}
Although our experiments are conducted in closed text-based benchmarks, stronger multi-turn agent training may lower the barrier to deploying autonomous language agents in open environments. If used without sufficient safeguards, such agents may execute unreliable action sequences, automate undesirable behavior, or waste external resources. Practical deployment should therefore pair GAGPO with sandboxing, permission control, and human oversight, especially in safety-critical settings.

\bibliography{custom}

\appendix

\section{Hyperparameter Sensitivity}
\label{sec:appendix_hparam}

We provide a sensitivity study for the two method-specific temporal hyperparameters in GAGPO, the discount factor $\gamma$ and the temporal propagation coefficient $\lambda$. We vary these parameters on ALFWorld with Qwen2.5-1.5B while keeping the remaining training pipeline unchanged, and report representative ALFWorld overall success results in Table~\ref{tab:hparam_gamma_lambda}. The default setting $(\gamma=0.95, \lambda=0.8)$ used in the main experiments yields the strongest overall performance, while nearby settings remain competitive, suggesting that GAGPO does not depend on a brittle single choice. In contrast, pushing $\gamma$ to 1.0 leads to a substantial degradation, indicating that overly long-horizon propagation amplifies estimation noise in practice. These trends are consistent with the design motivation of GAGPO: effective temporal credit assignment requires a balanced regime between myopic propagation and high-variance long-horizon recursion.

\begin{table}[t]
\centering
\caption{Sensitivity study for GAGPO temporal hyperparameters on ALFWorld with Qwen2.5-1.5B. We report representative ALFWorld overall success under selected $(\gamma, \lambda)$ configurations. The default setting used in the main paper is highlighted in bold.}
\label{tab:hparam_gamma_lambda}
\begin{tabular}{ccc}
\toprule
$\gamma$ & $\lambda$ & ALFWorld All \\
\midrule
0.80 & 0.70 & 90.8 \\
0.95 & 0.60 & 91.6 \\
0.95 & 0.70 & 92.4 \\
\textbf{0.95} & \textbf{0.80} & \textbf{93.5} \\
0.95 & 1.00 & 91.7 \\
1.00 & 0.80 & 82.7 \\
\bottomrule
\end{tabular}
\end{table}

\section{Training Configuration.}
For reproducibility, we summarize the key training settings used in our implementation in Table~\ref{tab:training_config}. Our implementation follows the official GiGPO/verl-agent training pipeline\footnote{\url{https://github.com/langfengQ/verl-agent}}. Unless otherwise specified, we keep all shared training and evaluation settings identical to the GiGPO baseline in our controlled experiments, including the environment setup, prompting format, rollout pipeline, evaluation protocol, optimizer type, batch construction, clipping coefficient, and KL regularization. The only differences are the proposed credit assignment mechanism and the method-specific temporal hyperparameters analyzed in Appendix~\ref{sec:appendix_hparam}.

\begin{table}[t]
\centering
\small
\begin{tabular}{ll}
\toprule
\textbf{Configuration} & \textbf{Value} \\
\midrule
Actor learning rate & $1\times 10^{-6}$ \\
KL loss coefficient & $0.01$ \\
KL estimator & Low-variance KL \\
Reward KL injection & Disabled \\
Invalid action penalty coefficient & $0.1$ \\
ALFWorld max prompt length & $2048$ \\
WebShop max prompt length & $4096$ \\
Max response length & $512$ \\
\bottomrule
\end{tabular}
\caption{Key training configurations used in the main experiments. Shared settings not listed here follow the official GiGPO/verl-agent training pipeline and are kept identical across controlled comparisons.}
\label{tab:training_config}
\end{table}

\section{Exact-Match Group Size Statistics}
\label{sec:appendix_group_dist}

Because both GAGPO and GiGPO construct rollout groups via exact textual state matching, one possible concern is that the gains of GAGPO might be explained by an easier grouping regime rather than by the proposed temporal estimator itself. Table~\ref{tab:group_size_stats} compares representative group-size statistics on ALFWorld with Qwen2.5-1.5B at training steps 60 and 120. The overall grouping regime remains broadly comparable across methods: singleton groups account for only a small fraction of sampled steps in both methods, mean group sizes remain in the same range, and medium-to-large groups still make up roughly half of the sampled steps. GAGPO exhibits slightly lower singleton mass and somewhat more large groups at the later stage, but it does not remove the singleton/small-group regime or induce a qualitatively different exact-match grouping pattern. These results support the interpretation that GAGPO's gains are driven mainly by improved temporal credit assignment under similar grouping conditions.

\begin{table}[t]
\centering
\caption{Representative exact-match group-size statistics on ALFWorld with Qwen2.5-1.5B. Percentages are measured over sampled steps.}
\label{tab:group_size_stats}
\resizebox{\linewidth}{!}{
\begin{tabular}{lccccc}
\toprule
Step & Method & Mean Size & Size 1 (\%) & Size $\leq 8$ (\%) & Size $\geq 16$ (\%) \\
\midrule
60 & GiGPO & 47.9 & 7.0 & 35.4 & 53.5 \\
60 & GAGPO & 43.1 & 4.9 & 30.8 & 53.6 \\
120 & GiGPO & 28.8 & 4.7 & 37.0 & 49.8 \\
120 & GAGPO & 33.0 & 3.8 & 27.2 & 58.3 \\
\bottomrule
\end{tabular}
}
\end{table}

\end{document}